\documentclass[10pt,twocolumn,letterpaper]{article}

\usepackage[pagenumbers]{cvpr} 

\usepackage{graphicx}
\usepackage{amsmath}
\usepackage{amssymb}
\usepackage{booktabs}
\usepackage{multirow}
\usepackage{algorithm, algorithmic}

\usepackage{stfloats}

\usepackage[pagebackref,breaklinks,colorlinks]{hyperref}

\usepackage[capitalize]{cleveref}
\crefname{section}{Sec.}{Secs.}
\Crefname{section}{Section}{Sections}
\Crefname{table}{Table}{Tables}
\crefname{table}{Tab.}{Tabs.}

\def\cvprPaperID{3894} 
\def\confName{CVPR}
\def\confYear{2023}

\begin{document}

\title{Self-similarity Driven Scale-invariant Learning for \\ Weakly Supervised Person Search}

\author{Benzhi Wang$^{1,2}$, Yang Yang$^{1,2}$, Jinlin Wu$^{1,2,4}$, Guo-jun Qi$^{3}$, Zhen Lei$^{1,2,4}$\\
$^{1}$CBSR \& NLPR, Institute of Automation, Chinese Academy of Sciences\\
$^{2}$University of Chinese Academy of Sciences\\
$^{3}$MAchine Perception and LEarning (MAPLE) Lab, the
University of Central Florida\\
$^{4}$Centre for Artificial Intelligence and Robotics, Hong Kong Institute of Science \& Innovation,\\
Chinese Academy of Sciences\\
{\tt\small wangbenzhi2021@ia.ac.cn, \{yang.yang, jinlin.wu, zlei\}@nlpr.ia.ac.cn, guojunq@gmail.com}
}
\maketitle

\begin{abstract}
Weakly supervised person search aims to jointly detect and match persons with only bounding box annotations. Existing approaches typically focus on improving the features by exploring relations of persons. However, scale variation problem is a more severe obstacle and under-studied that a person often owns images with different scales (resolutions). On the one hand, 
small-scale images contain less information of a person, thus affecting the accuracy of the generated pseudo labels. On the other hand, the similarity of cross-scale images is often smaller than that of images with the same scale for a person, which will increase the difficulty of matching. In this paper, we address this problem by proposing a novel one-step framework, named Self-similarity driven Scale-invariant Learning (SSL). Scale invariance can be explored based on the self-similarity prior that it shows the same statistical properties of an image at different scales. To this end, we introduce a Multi-scale Exemplar Branch to guide the network in concentrating on the foreground and learning scale-invariant features by hard exemplars mining. To enhance the discriminative power of the features in an unsupervised manner, we introduce a dynamic multi-label prediction which progressively seeks true labels for training. It is adaptable to different types of unlabeled data and serves as a compensation for clustering based strategy. Experiments on PRW and CUHK-SYSU databases demonstrate the effectiveness of our method.
\end{abstract}

\section{Introduction}
\label{sec:intro}
Recent years have witnessed remarkable success of person search which is to match persons existed in real-world scene images. It is often taken as a joint task consisting of person detection~\cite{liu2019high-pedestrian-detection1,pang2019mask-pedestrian-detection2,zhang2018occlusion-pedestrian-detection3} and re-identification (re-id) \cite{sun2018beyond-reid-1,yang2014salient-reid-2,wang2020cross-reid-3}. To achieve high performance, existing methods are commonly trained in a fully supervised setting ~\cite{chen2018person-mask-MGTS,cao2022pstr-PSTR,yan2021anchor-AlignPS,yu2022cascade-COAT,han2021end-AGWF,lee2022oimnet++,li2021sequential-SeqNet,chen2020norm-NAE} where the bounding boxes and identity labels are required. However, it is time-consuming and labor-intensive to annotate both of them in a large-scale dataset, which encourages some researchers to embark on reducing the supervision. 

\begin{figure}
    \centering
   \includegraphics[width=\linewidth]{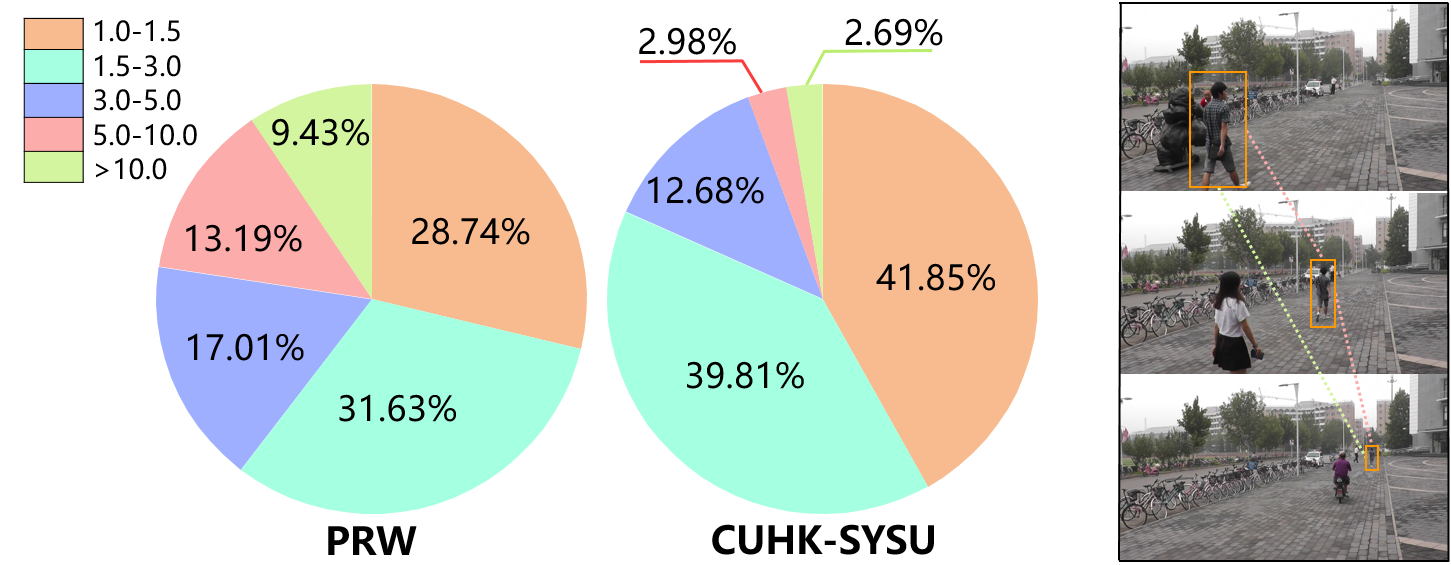}
    \caption{The scale variation of the same person on PRW and CUHK-SYSU datasets.}
    \label{fig:scale-variation}
\end{figure}

Considering that it is much easier to annotate bounding boxes than person identities, we dedicate this paper to weakly supervised person search which only needs bounding box annotations. Intuitively, we can address it with a supervised detection model and an unsupervised re-id model \cite{wu2019unsupervised-jinlin1,yu2019unsupervised-soft-multilabel, fan2018unsupervised-reid-clustering-and-finetuning, ge2020self-paced,cho2022part-part-pseudo-labels} independently. To be specific, we first train a detector to crop person images and then apply an unsupervised re-id model for matching, which is regarded as a two-step person search model. Nevertheless, one of the major drawbacks of such two-step methods is low-efficiency, i.e., it is of high computational cost with two network parameters during training and inconvenient for test. In contrast, one-step methods can be trained and tested more effectively and efficiently \cite{han2021weakly-R-SiamNet,yan2022exploring-CGPS}. Han et al. \cite{han2021weakly-R-SiamNet} use a Region Siamese Network to learn consistent features by examining relations between auto-cropped images and manually cropped ones. Yan et al. \cite{yan2022exploring-CGPS} learn discriminative features by exploring the visual context clues. According to the learned features, both of them generate pseudo labels via clustering. Although promising results are achieved, they fail to take into account the scale variation problem that a person often owns images with different scales (resolutions) because the same person is captured at different distances and camera views. As shown in \cref{fig:scale-variation}, the images of a person from PRW and CUHK-SYSU datasets have large variation in scale. Since it is unable to resize the input images to a fixed scale for one-step methods, the existing scale variation problem will further affect the procedure of the pseudo label prediction and the subsequent person matching.

In this paper, we propose a novel Self-similarity driven Scale-invariant Learning (SSL) weakly supervised person search framework to solve the scale variation problem. It consists of two branches: Main Branch and Multi-scale Exemplar Branch. The former branch takes the scene image as the input and applies a detector to extract RoI features for each person. Motivated by the self-similarity prior \cite{Huang2015Single} that one subject is similar at different scales, we design the latter branch which is served as a guidance of the former to learn body-aware and scale-invariant features. Firstly, we crop the foreground of person images by using the given bounding boxes and generated binary masks. Each cropped image is regarded as an exemplar. Secondly, we resize each of the exemplars to several fixed scales. At last, we formulate a scale-invariant loss by hard exemplar mining. Note that RoI features in Main Branch focuses on differentiating a person from background but are not able to distinguish different persons. Guided by Multi-scale Exemplar Branch, we can enable Main Branch to learn both scale-invariant and discriminative features. To generate reliable pseudo labels, we introduce a dynamic threshold for multi-label learning. It can find true labels progressively and be adaptable to different datasets, which serves as an compensation for cluster-level prediction \cite{ge2020self-paced}. Finally, we integrate the scale-invariant loss, multi-label classification loss and contrastive learning loss together and optimize them 
jointly.     

Our contributions are summarized as follows:

\begin{itemize}
    \item We propose a novel end-to-end Self-similarity driven Scale-invariant Learning framework to solve the task of weakly supervised person search. It bridges the gap between person detection and re-id by using a multi-scale exemplar branch as a guidance.
    \item We design a scale-invariant loss to solve the scale variation problem and a dynamic multi-label learning which is adaptable to different datasets.    
    \item We confirm the efficacy of the proposed method by achieving state-of-the-art performance on PRW and CUHK-SYSU datasets. 
\end{itemize}
\section{Related Work}

\textbf{Person Search.} Nowadays, person search has attracted increasing attention because of its wide application in real-world environment. Its task is to retrieve a specific person from a gallery set of scene images. It can be seen as an extension of re-id task by adding a person detection task. 

Existing methods addressing this task can be classified to two manners: one-step~\cite{xiao2017joint-OIM,yan2021anchor-AlignPS,cao2022pstr-PSTR, yu2022cascade-COAT,han2021decoupled} and two-step ~\cite{han2019-RDLR, dong2020instance-IGPN,chen2018person-mask-MGTS} methods. One-step methods tackle person detection and re-id simultaneously. The work~\cite{xiao2017joint-OIM} proposes the first one-step person search approach based on deep learning. It provides a practical baseline and proposes Online Instance Matching(OIM), which is still used in recent works. Yan \etal~\cite{yan2021anchor-AlignPS} introduce an anchor-free framework into person search task and tackle the misalignment issues at different levels. Dong \etal~\cite{dong2020bi-BINet} propose a bi-directional interaction network and use the cropped image to alleviate the influence of the context information. In contrast, two-step methods process person detection and re-id separately, which alleviates the conflict between them~\cite{han2021end-AGWF}. Chen \etal~\cite{chen2018person-mask-MGTS} introduce an attention mechanism to obtain more discriminative re-id features by modeling the foreground and the original image patches. Wang \etal~\cite{wang2020tcts-TCTS} propose an identity-guided query detector to filter out the low-confidence proposals. 

Due to the high cost of obtaining the annotated data, Li \etal~\cite{li2022domain} propose a domain adaptive method. In this setting, the model is trained on the labeled source domain and transferred to the unlabeled target domain. Recent years, Han \etal~\cite{han2021weakly-R-SiamNet} and Yan \etal~\cite{ yan2022exploring-CGPS} propose the weakly jsupervised person search methods, which only needs the bounding box annotations. Due to the absence of person ID annotations in the weakly supervised setting, we need to generate pseudo labels to guide the training procedure. The quality of the pseudo labels has a significant impact on the performance. Thus, how to generate reliable pseudo labels is an important issue of the weakly supervised person search.

\textbf{Unsupervised re-id.} Due to the limitation of the annotated data, fully supervised re-id has poor scalability. Thus, lots of unsupervised re-id are proposed, which try to generate reliable yet valid pseudo labels for unsupervised learning. Some of them consider the global data relationship and apply unsupervised cluster algorithm to generate pseudo labels. For example, Fan \etal~\cite{fan2018unsupervised-reid-clustering-and-finetuning} propose an iterative clustering and fine-tuning method for unsupervised re-id. Ge \etal~\cite{ge2020self-paced} use the self-paced strategy to generate pseudo labels based on the clustering method. Cho \etal~\cite{cho2022part-part-pseudo-labels} propose a pseudo labels refinement strategy based on the part feature information. Although cluster-level methods have made great progress, the within-class noisy introduced by cluster algorithm still limits the further improvement. 

To solve this issue, another methods introduce fine-grid instance-level pseudo labels as the supervision for unsupervised learning. For example, zhong \etal~\cite{zhong2019invariance} propose an instance exemplar memory learning scheme that considers three invariant cues as the instance-level supervision, including exemplar-invariance, camera invariance and neighborhood-invariance. Lv \etal~\cite{lv2018unsupervised} look for underlying positive pairs on the instance memory with the guiding of spatio-temporal. information. Wang \etal~\cite{wang2020unsupervised} consider the instance visual similarity and the instance cycle consistency as the supervision. Lin \etal~\cite{lin2020soften-multi-label} regard each training image as a single class and train the model with the soften label distribution.

\textbf{Multi-scale Matching Problem.} Person search suffers from the multi-scale matching problem because of the scale variations in scene images. Lan \etal~\cite{lan2018person-multi-scale-matching-CLSA} propose a two-step method with knowledge distillation to alleviate this problem. Unlike the fully supervised setting, due to the absence of the ID annotations, it is much harder to learn a consistent feature for the same person appears in various scales. In this paper, we propose the Self-similarity driven Scale-invariant Learning to improve the feature consistency among different scales in the weakly supervised setting.

\begin{figure*}[t]
  \centering
   \includegraphics[width=1\linewidth]{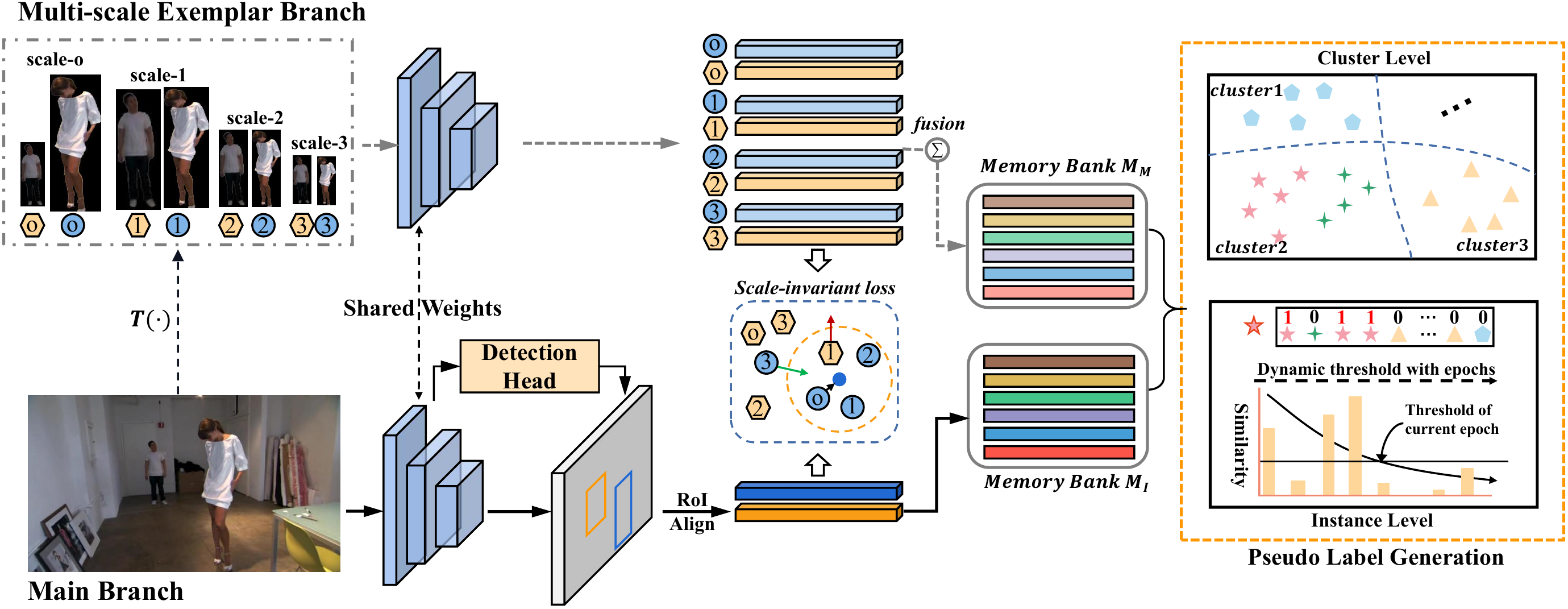}
   \caption{Details of our SSL for weakly supervised person search. The SSL consists of the multi-scale exemplar branch, main branch and two extra memory banks. The Main branch takes the scene image as input, which is utilized to detect persons and extract their re-id features. Given the multi-scale images (original scale and three different scales for example) corresponding to the persons in the scene image, the multi-scale exemplar branch takes them as input and obtains the multi-scale features. We conduct the scale-invariant loss (SL) between instance features and multi-scale features to learn scale-invariant features. 
   We also adopt our dynamic threshold multi-label classification strategy and clustering algorithm to obtain reliable yet valid pseudo labels as the supervision for unsupervised learning. 
   }
   \label{fig:framework}
\end{figure*}

\section{Proposed Method}
In this section, we first introduce the overall framework in \cref{sec: overall-framework}, then describe the scale-invariant learning in \cref{sec: scale-invariant-feature-learning}. A reliable pseudo label generation is detailed in \cref{sec: Dynamic Multi-label Learning} and the training and inference procedure is finally explained in \cref{sec: Training and Inference}.

\subsection{Framework Overview}
\label{sec: overall-framework}
Different from the fully supervised person search, only the bounding box annotations are accessible in the weakly supervised setting. Firstly, we propose the scale-invariant loss to address the scale variance problem by hard exemplar mining. In addition, because the existing cluster based methods may incur noisy pseudo labels, we further propose a dynamic threshold based method to obtain pseudo labels, which is jointly used with cluster based method. 

The general pipeline of the framework is illustrated in \cref{fig:framework}. Our detection part is based on Faster R -CNN~\cite{ren2015faster-fasterRCNN}, a widely used object detection baseline.
As aforementioned, scale variation
problem is a severe obstacle and will
further affect the procedure of the pseudo label prediction and the subsequent person matching . To address it, we propose the SSL that consists of multi-scale exemplar branch and main branch. The main branch locates the persons firstly, and extracts the RoI features by the RoI-Align layer with the localization information. The multi-scale exemplar branch is used to obtain the features of the same person with different scales. Specifically, the multi-scale cropped images with background filtering~\cite{li2020self-correction-parsing} and scene images are fed into the two branches for scale-invariant feature learning. 

To enhance the reliability of the pseudo labels, we propose a dynamic threshold based method to obtain the instance level pseudo labels. At the same time, the DBSCAN based method is also used to obtain complementary cluster level pseudo labels. 

\subsection{Scale-invariant Learning}
\label{sec: scale-invariant-feature-learning}

In this section, we adopt a Scale Augmentation strategy to obtain multi-scale exemplars and propose a Scale-invariant Learning to learn scale-invariant features by hard exemplar mining.
\vspace{1mm}

\textbf{Scale Augmentation}. Given a scene image $I$, we obtain the cropped image of the $i$-th person $x^i_o$ with the given localization annotation $gt_i$.

Then, we apply a binary mask~\cite{li2020self-correction-parsing} filtering the background to obtain person's emphasized region:
\begin{equation}
    x^i_s \longleftarrow \mathcal{T}(x^i_o \odot \rm mask, s),
\end{equation}
where $\odot$ means pixel-wise multiplication, and $\mathcal{T}_s$ is the bilinear interpolation function that transforms the masked image to the corresponding scale $s$. 
\vspace{1mm}

\begin{figure}[t]
  \centering
  \includegraphics[width=0.9\linewidth]{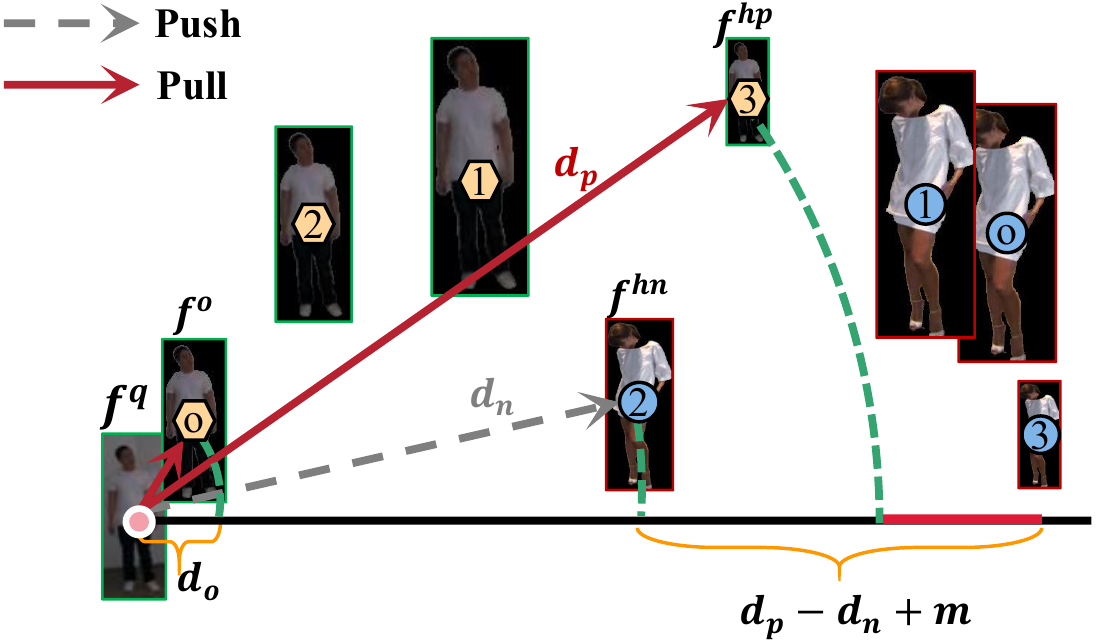}

  \caption{Illustration of the scale-invariant loss. Given the query person, we obtain the corresponding multi-scale features via our multi-scale exemplar branch. $m$ is the distance margin, $d_o$ denotes the distance between the query feature$f^q$ and its original scale feature $f^o$, $d_p$ denotes the distance between the $f^q$ and its hardest distinguish positive scale feature $f^{hp}$. The $d_n$ denotes the distance between the $f^q$ and its most confusing negative scale feature $f^{hn}$.}
  \label{fig:scale-invariant-loss}
\end{figure}

\vspace{1mm}
\textbf{Scale-invariant Loss.} When we have obtained multi-scale exemplars by scale augmentation, we use them to learn the scale-invariant features in the guidance of multi-scale exemplar branch which takes the exemplars as the inputs and aims to extract multi-scale features. At the same time, our main branch takes the scene image as input, aiming to detect persons and extract corresponding re-id features.

In a scene image, assuming there are $\mathbf{P}$ persons, we can get corresponding cropped images by the bounding box annotations, and we augment each bounding box to $\mathbf{K}$ different scales. Thus, we totally obtain $\mathbf{P(K+1)}$ cropped images with different scales. The scale-invariant loss can be formulated as follows:
\begin{small}
\begin{equation}
        L_{scale}^b = \frac{1}{P}\sum_{i=1}^{P}([\ m+\max\limits_{s=1\dots K}D_{s}^{i\rightarrow{i}}
        -\min\limits_{
        \substack{j=1\dots P\\  s=1\dots K+1\\  j\neq i}}
        D_{s}^{i\rightarrow{j}}]_+ + \gamma D_{o}^{i\rightarrow{i}}),
    \label{eq:scale-invariant-loss}
\end{equation}
\end{small}
with
\begin{equation}
    \begin{split}
    D_{s}^{i\rightarrow{i}}&=D(f_i, f_i^s),\\
    D_{s}^{i\rightarrow{j}}&=D(f_i, f_j^s),\\
    D_{o}^{i\rightarrow{i}} &= D(f_i, f_i^o),
    \end{split}
\end{equation}
\begin{equation}
    L_{scale} = \frac{1}{B}\sum_{b=1}^B L_{scale}^b,
\end{equation}
where $B$ denotes the batch size, $D(\cdot)$ measures the squared euclidean distance between two features, $f_i$ stand for the instance feature extracted from the main branch. $f_j^s$ and $f_i^s$ stand for the multi-scale features of the $i$-th and $j$-th persons, respectively. And $f_i^o$ means for the original scale feature of the $i$-th person. $m$ denotes the distance margin and $\gamma$ denotes the regularization factor. Furthermore, the obtained features are processed by $l2$-norm.

An intuitive illustration of $L_{scale}$ is presented in \cref{fig:scale-invariant-loss}. $L_{scale}$ tries to select the most difficult positive and negative ones from multi-scale exemplars for a query image. It learns the scale-invariant features by decreasing the distance with the corresponding hardest positive exemplar and increasing the distance with the corresponding hardest negative exemplar. Meanwhile, it is notably that the original scale cropped image could be aligned better with the instance feature. Thus, we additionally constrain the distance between the instance feature and its corresponding original scale feature to make the model focus on the foreground information and extract body-aware features.

\subsection{Dynamic Multi-label Learning}
\label{sec: Dynamic Multi-label Learning}
In weakly supervised settings, we do not have the ID annotations of each person. Thus, it is very important to predict pseudo labels and its quality will affect the subsequent training for discriminative re-id features. 
\vspace{1mm}

As shown in \cref{fig:framework}, we maintain two extra memory banks $\mathcal{M}_{I}\in\mathbb{R}^{N\times d}$ and $\mathcal{M}_{M}\in\mathbb{R}^{N\times d}$ to store the features extracted from the main branch and multi-scale exeplar branch, separately. Where $N$ is the number of samples in the training data set and $d$ is the feature dimension.
For the latter branch, we extract features $[f_i^{o}, f_i^{1},f_i^{2}, f_i^{3}]$ corresponding to scale-o, scale-1, scale-2, and scale-3, and obtain the average feature $f_{i}^h=\mathrm{mean}(f_i^{o}, f_i^{1},f_i^{2}, f_i^{3})$. For the main branch, we extract the instance feature $f_i$ from the scene image.
After each training iteration, $\mathcal{M}_{M}$ and $\mathcal{M}_{I}$ are updated as:
\begin{equation}
\begin{split}
        \mathcal{M}_{M}[i] &= \lambda \cdot f_{i}^{h} + (1-\lambda) \cdot \mathcal{M}_{M}[i],\\
        \mathcal{M}_{I}[i] &= \lambda \cdot f_{i} + (1-\lambda) \cdot \mathcal{M}_{I}[i],
    \label{eq:Memoryupdate}
\end{split}
\end{equation} 
where $\lambda$ is the momentum factor and set to 0.8 in our experiments.   

To obtain reliable pseudo labels, we use the multi-scale features to generate pseudo labels for clustering and multi-label classification. The multi-scale features are obtained as follows:
\begin{equation}
   \mathcal{M} = \textrm{mean}(\mathcal{M}_{M},\mathcal{M}_{I}).
    \label{con:scale-enhancement-feature}
\end{equation} 

 \vspace{1mm}
\textbf{Dynamic Pseudo Label Prediction.} Suppose we have a training set $\mathcal{X}$ with $N$ samples, we treat the pseudo label generation as an N-classes multi-label classification problem. In other words, the $i$-th person has an N-dim two-valued label $Y_i=[Y_{i1},Y_{i2} \dots, Y_{iN}]$. The label of the $i$-th person can be predicted based on the similarity between its feature $f_i$ and the features of others. Based on the $\mathcal{M}$, the similarity matrix can be obtained as follows:
\begin{equation}
    S=\mathcal{M}\mathcal{M}^T = \left[
    \begin{array}{ccc}
        s_{11}, &\dots , &s_{1N}\\
        \vdots &\ddots &\vdots\\
        s_{N1},&\dots , &s_{NN}
    \end{array}
    \right], \label{con:similarity}
\end{equation}
and with it, we can get two-valued labels $Y$ matrix with a threshold $t$:
\begin{equation}
    Y_{i,j}= \left\{
    \begin{array}{cc}
        1 & s_{ij} \geq t \\
        0 & s_{ij} < t
    \end{array} 
    \right..\label{con:label_generation}
\end{equation}
\label{eq:multi-label classification without any whistles}

However, the multi-label classification method is sensitive to the threshold. An unsuitable threshold can seriously affect the quality of label generation, i.e., the low threshold will introduce a lot of noisy samples, while the high threshold omits some hard positive samples. 
Thus, we further adopt an exponential dynamic threshold to generate more reliable pseudo labels. That is,
\begin{equation}
    t = t_s + \alpha\cdot e^{\beta \cdot e},
\end{equation}
where $t_s$ is the initial threshold, $\alpha$ and $\beta$ are the ratio factors, and $e$ stands for current epoch number. So far, we can use the dynamic threshold to get the label vector for each person at each iteration by \cref{con:label_generation}.

We define the positive label set of the $i$-th person $P_i$ and negative label set $N_i$. To make the pseudo label more reliable,  we further process the label based on the hypothesis: \textit{persons in the same image can not be the same person.} For the $i$-th person, we can get its similarity vector $S_i$ by \cref{con:similarity}. We sort the $S_i$ in descending order and get the sorted index:
\begin{equation}
     {SI}_i = \mathop{\arg \mathrm{sort}(s_{ij})}_{j\in P_i} \ \ \ \   \mathrm{w.r.t.}, 1\leq j \leq n.
\end{equation}

Then, we traverse the label $\mathcal{Y}_i$ by the $SI_i$. If the $j$-th person is predicted 
to be the same person with the $i$-th person, i.e., $\mathcal{Y}_{i,j}=1$. We consider the other persons belong to the same image with the $j$-th person can not have the same ID with the $i$-th person, and set these labels to 0. 

Besides, for the cluster level, based on the $\mathcal{M}$, we adopt DBSCAN with self-paced strategy~\cite{ge2020self-paced} to generate cluster-level pseudo labels.

\textbf{$\mathcal{M}_M$ and $\mathcal{M}_I$ based re-id feature learning.} As aforementioned, we use the $\mathcal{M}$ to generate reliable pseudo labels and calculate the loss function on the two branches with $\mathcal{M}_{I}$ and $\mathcal{M}_{M}$, separately. The instance level multi-label learning loss function can be formulated as:
\begin{small}
    \begin{equation}
\begin{split}
        L_{ml}(\mathcal{M}^*,f^*) = \sum\limits_{i=1}^q[\frac{\delta}{|P_i|}\sum\limits_{p\in P_i}||\mathcal{M}^*[p]^{T}\times f^* + (-1)^{Y_{i,p}}||^2 + \\
            \frac{1}{|N_i|}\sum\limits_{v\in N_i}||\mathcal{M}^*[v]^{T}\times f^* + (-1)^{Y_{i,v}}||^2],
\end{split}
\end{equation}
\end{small}
where $\mathcal{M}^*\in \{\mathcal{M}_I, \mathcal{M}_M\}$, $f^* \in \{f_i, f_i^h\}$, $q$ is the number of persons in a mini-batch and $\delta$ is used as a balance factor of the loss. The total dynamic multi-label learning loss can be formulated as follows: 
\begin{equation}
    L_{DML} = L_{ml}(\mathcal{M}_I,f_i) + L_{ml}(\mathcal{M}_M,f_i^h).
\end{equation}

\subsection{Training and Inference}
\label{sec: Training and Inference}

In general, our SSL is trained in an end-to-end manner by using the following loss function:
\begin{equation}
    L = L_{scale}+(L_{cluster} +L_{DML})+L_{det},
\end{equation}
where $L_{det}$ stands for the detection loss used in SeqNet~\cite{li2021sequential-SeqNet} and $L_{cluster}$ denotes to the contrastive learning loss used in CGPS~\cite{yan2022exploring-CGPS}.
\label{sec:traing procedure}

In the inference phase, we only use the main branch to detect the persons and extract the re-id features which are further used to compute their similarity score.
\section{Experiments}

\subsection{Datasets and Settings}
\textbf{CUHK-SYSU}~\cite{xiao2017joint-OIM} is one of the largest public datasets for person search. It contains 18,184 images, including 12,490 frames from street scenes and 5,694 frames captured from movie snapshots. CUHK-SYSU provides 8,432 annotated identities and 96,143 annotated bounding boxes in total, where the training set contains 11,206 images and 5,532 identities, and the test set contains 6,978 images and 2,900 query persons. CUHK-SYSU also provides a set of evaluation protocols with gallery sizes from 50 to 4000. In this paper, we report the results with the default 100 gallery size.

\textbf{PRW}~\cite{zheng2017person-PRW} is collected in a university campus by six cameras. The images are annotated every 25 frames from a 10 hours video. It contains 11,816 frames with 43,110 annotated bounding boxes. The training set contains 5,401 images and 482 identifies, and the test set contains 6,112 images and 2,507 queries with 450 identities.

\textbf{Evaluation Protocol.} We adopt the Cumulative Matching Characteristic (CMC), and the mean Averaged Precision (mAP) to evaluate the performance for person search. We also adopt recall and average precision to evaluate person detection performance.

\subsection{Implementation Details}

 We adopt ResNet50~\cite{he2016resnet} pre-trained on ImageNet~\cite{deng2009imagenet} as our backbone. We set the batch size to 2 and adopt the stochastic gradient descent (SGD) algorithm to optimize the model for 26 epochs. The initial learning rate is 0.001 and is reduced by a factor of 10 at 16 and 22 epochs. 
We set the momentum and weight decay to 0.9 and $5\times10^{-4}$, respectively. We set the hyperparameters $m=0.3,\ t_s=0.6,\ \alpha=0.1,\ \beta=-0.1$ and $\gamma=0.05$. We employ DBSCAN~\cite{ester1996DBSCAN} with self-paced learning strategy~\cite{ge2020self-paced} as the basic clustering method, and the hyper-parameters are the same as~\cite{yan2022exploring-CGPS}. 
In our experiments, the scale-1, scale-2, scale-3 are set to 112$\times$48, 224$\times$96 and 448$\times$192, respectively. 
For inference, we resize the images to a fixed size of 1500 $\times$ 900 pixels. Furthermore, We use PyTorch to implement our model, and run all the experiments on an NVIDIA Tesla V100 GPU.

\subsection{Ablation Study}

\begin{table}[htbp]
  \centering

 \resizebox{\linewidth}{!}{
    \begin{tabular}{ccccc|cc}
    \toprule
    \multicolumn{1}{c}{\multirow{2}[4]{*}{Baseline}} & \multicolumn{1}{c}{\multirow{2}[4]{*}{ DML }} & \multicolumn{1}{c}{\multirow{2}[4]{*}{SL}} & \multicolumn{2}{c|}{PRW} & \multicolumn{2}{c}{CUHK-SYSU} \\
\cmidrule{4-7}          &       &       & \multicolumn{1}{c}{mAP} & \multicolumn{1}{c|}{top-1} & \multicolumn{1}{c}{mAP} & \multicolumn{1}{c}{top-1} \\
    \midrule
       \midrule
         \checkmark &       &       &  18.8     &   67.0    &    80.7   & 82.5 \\
         \checkmark &    \checkmark  &   &  20.8     &   70.8    &    84.5   &   86.3     \\
       \checkmark   &       &     \checkmark  &  26.2     &   76.4    &   86.2    & 87.4 \\
        \checkmark  &    \checkmark   &    \checkmark   &  30.7     &  80.6     &   87.4    & 88.5 \\
    \bottomrule
    \end{tabular}%
    }
      \caption{Ablation study on the two key components of our approach. We report the mAP(\%) and top-1 accuracy(\%) on PRW and CUHK-SYSU. DML denotes dynamic multi-label learning and SL means scale-invariant learning.}
  \label{tab:ablation study}%
\end{table}%

\textbf{Baseline.} We adopt a classical two-stage Faster R-CNN detector as our baseline model. Following SeqNet~\cite{li2021sequential-SeqNet}, we adopt two RPN structure to obtain more quality proposals. Furthermore, we adopt DBSCAN~\cite{ester1996DBSCAN} with self-paced strategy~\cite{ge2020self-paced} to generate pseudo labels and optimize the model with the contrastive learning loss in \cite{yan2022exploring-CGPS}.

\vspace{1mm}
\textbf{Effectiveness of Each Component.} We analyze the effectiveness of our SSL framework, and report the results in \cref{tab:ablation study}, where DML denotes dynamic multi-label learning and SL means scale-invariant learning. 

Firstly, we can see that the baseline model achieves 18.8\% mAP and 67.0\% top-1 on PRW. With the SL, baseline obviously improves the mAP and top-1 by 7.4\% and 9.4\% on PRW, respectively. This improvement indicates that the proposed SSL is effectiveness to handle pedestrian scale variations. Secondly, we can observe that DML improves baseline model by 2.0\% in mAP and 3.8\% in top-1 on PRW dataset, and DML further improves baseline + SL by 4.5\%/4.2\% in mAP/top-1 and improves baseline + SL by 1.2\%/1.1\% in mAP/top-1 on CUHK-SYSU. This improvement illustrates the effectiveness of our dynamic multi-label classification strategy for unsupervised learning. 
\begin{table}[h]
 \centering

   \begin{tabular}{llcc|cc}
   \toprule
   \multicolumn{2}{c}{\multirow{2}[4]{*}{Methods}} & \multicolumn{2}{c|}{ PRW } & \multicolumn{2}{c}{ CUHK-SYSU } \\
\cmidrule{3-6}    \multicolumn{2}{c}{} & \multicolumn{1}{c|}{ mAP } &  top-1  & \multicolumn{1}{c|}{mAP} & top-1 \\
   \midrule
   \midrule
 \multicolumn{2}{l}{SSL w/ Original Scale}  & 23.9     & 75.6     & 84.3     & 86.1 \\
      \multicolumn{2}{l}{SSL w/ One Scale} & 24.3     & 75.9     & 84.8     & 86.3 \\
      \multicolumn{2}{l}{SSL w/ Multi-Scale} & 27.1     & 77.4    & 86.1     & 87.9 \\
   \multicolumn{2}{l}{SSL w/ Multi-Scale\dag} & 28.6     & 79.6     & 86.9     & 88.2 \\
   \bottomrule
   \end{tabular}%
    \caption{Comparison of different scale settings on CUHK-SYSU and PRW. $\dag$ means filtering the background of the cropped images. W/ means combined with.}
 \label{tab:ContextAndScale}%
\end{table}%

\vspace{1mm}
\textbf{Effectiveness of Scale-invariant Loss.} In the \cref{sec: Dynamic Multi-label Learning}, we introduce using the multi-scale features to generate more reliable pseudo labels. For a fair comparison, we only use the memory bank $M_I$ to generate pseudo labels in our experiments. The results are reported in \cref{tab:ContextAndScale}.
In the method \textit{SSL w/ Original Scale}, we only use the original size of the cropped person images, which is obtained by bounding box annotations directly. In the method \textit{SSL w/ One Scale}, we resize the cropped persons images to 224$\times$96\ pixels. We observe that the \textit{One Scale} method just surpasses the \textit{Original Scale} method by 0.4\% and 0.3\% in mAP and top-1. The \textit{SSL w/ Multi-Scale} method significantly improves the performance, which achieves 27.1\% in mAP and 77.4\% in top-1. $\dag$ means filtering the background with the method in \cref{sec: scale-invariant-feature-learning}, which makes the model concentrate more on the foreground information and obtains more discriminative features. Filtering the background information further improves the performance by 1.5\% and 2.2\% in terms of mAP and top-1.

\begin{table}[t]
 \centering
\setlength{\tabcolsep}{3mm}
  \begin{tabular}{llcc|cc}

   \toprule
   \multicolumn{2}{c}{\multirow{2}[4]{*}{Methods}} & \multicolumn{2}{c|}{ PRW } & \multicolumn{2}{c}{ CUHK-SYSU } \\
\cmidrule{3-6}    \multicolumn{2}{c}{} & \multicolumn{1}{c|}{ mAP } &  top-1  & \multicolumn{1}{c|}{mAP} & top-1 \\
   \midrule
   \midrule

\multicolumn{2}{l}{ML} &  29.2    & 79.0    &   84.2 & 85.0  \\

\multicolumn{2}{l}{Ours} & 30.7     & 80.6     & 87.4     & 88.5 \\
   \bottomrule

   \end{tabular}%
  \caption{Effect of the reliable pseudo label. ML denotes using multi-label classification without bells and whistles.}
 \label{tab:dynamic threshold multi-label classification}%
\end{table}%

\begin{figure}[h]
  \centering
  \begin{subfigure}{0.6\linewidth}
\includegraphics[width=1\linewidth]{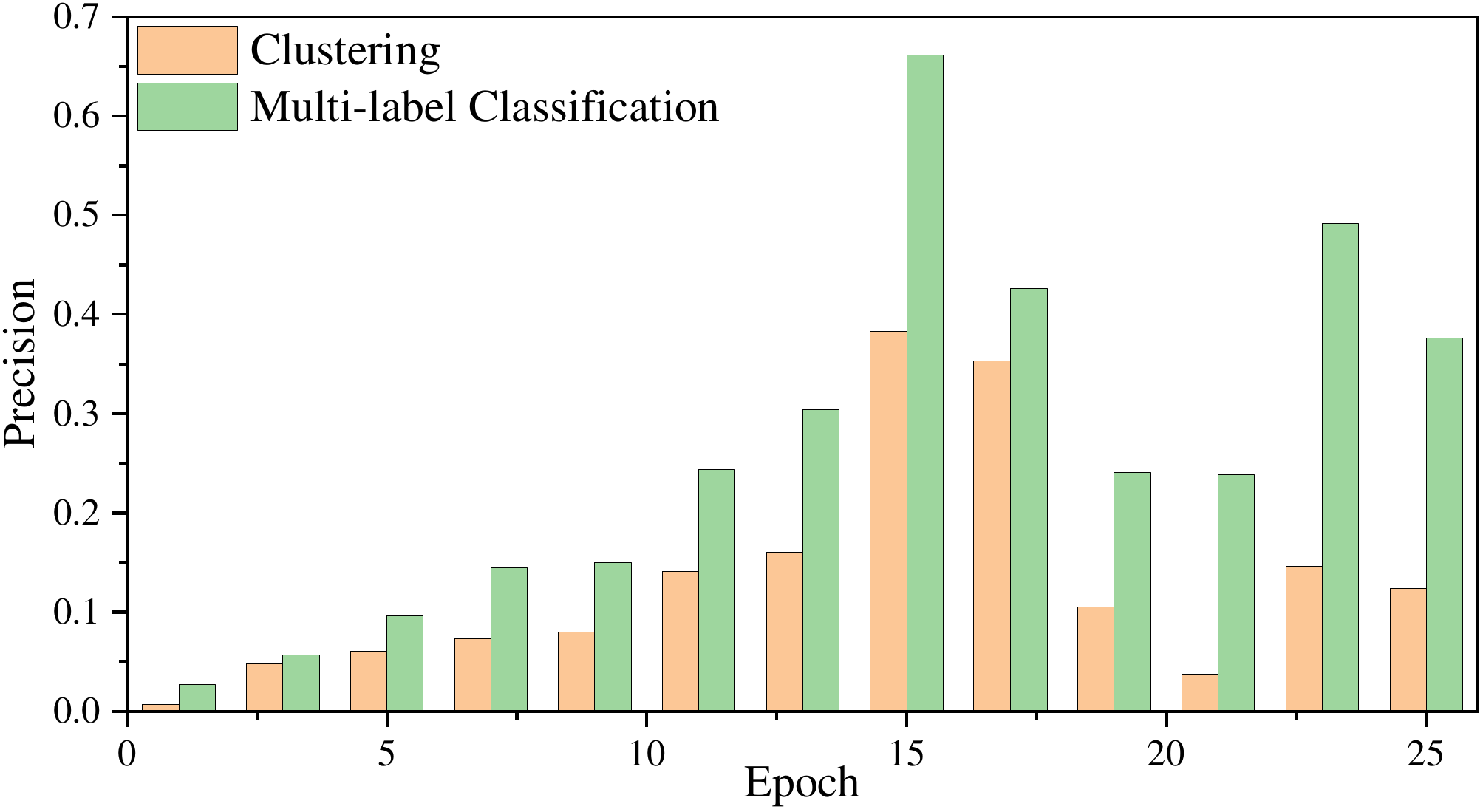}
    \caption{precision}
    \label{fig:short-a}
  \end{subfigure}
    \begin{subfigure}{0.6\linewidth}
\includegraphics[width=1\linewidth]{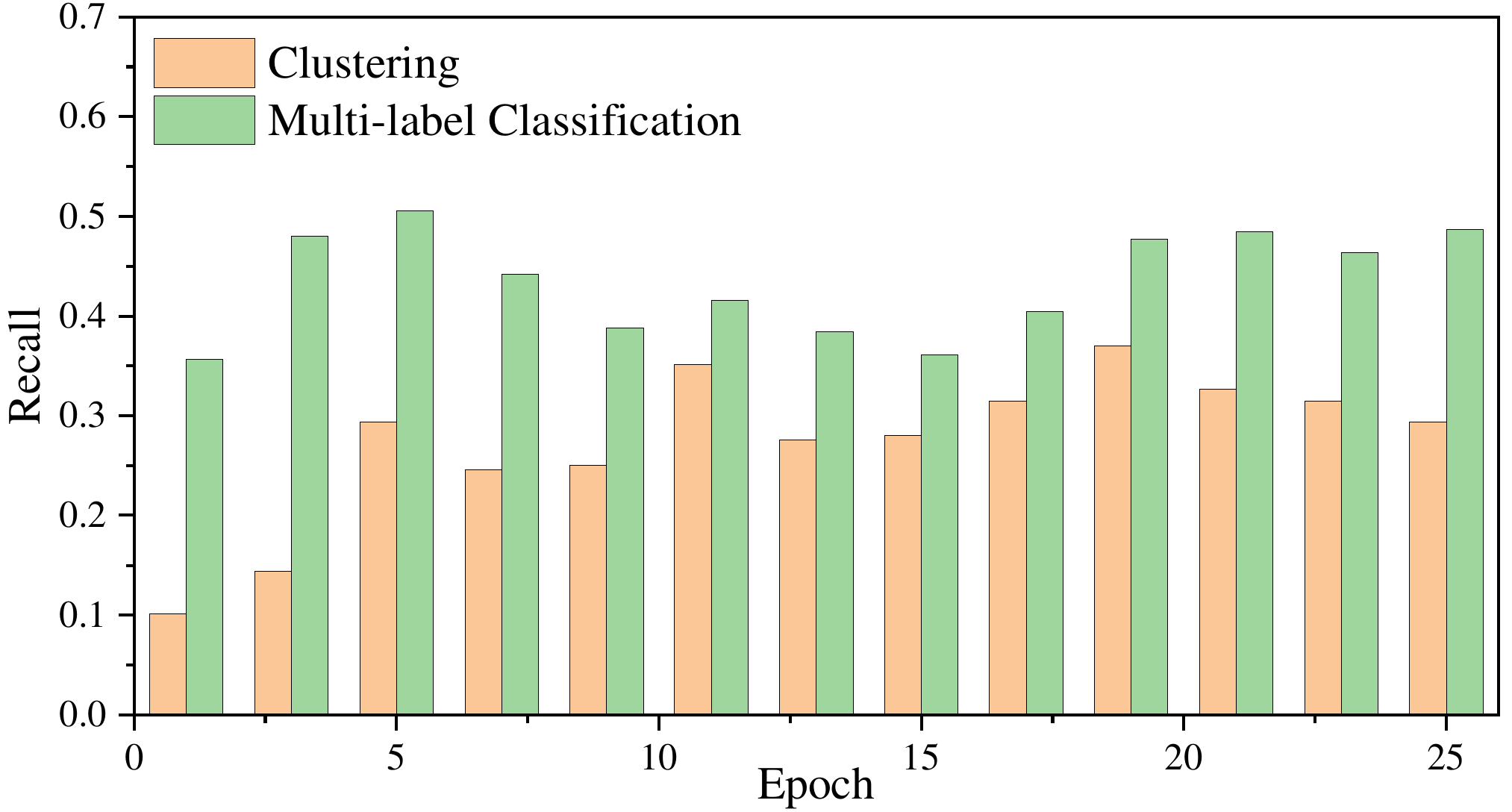}
    \caption{recall}
    \label{fig:short-b}
  \end{subfigure}
  \caption{Evaluation on recall and precision of clustering method and multi-label classification.}
  \label{fig:precision and recall}
\end{figure}
\textbf{Effectiveness of Pseudo Label Generation Method}. To illustrate the effectiveness of DML, we count the precision and recall of the  pseudo labels, and report the statistics result in \cref{fig:precision and recall}. Compared with clustering method, the higher precision indicates that DML is more reliable, while the higher recall indicates that DML is more valid. This is because that cluster method (i.e., DBSCAN) introduces within-class noisy. Our DML can alleviate this issue. 
Additionally, we shows the limitation of the multi-label classification with a fixed threshold in ~\cref{tab:dynamic threshold multi-label classification}, where ML denotes generating the pseudo multi-labels with a fixed threshold. It shows that the ML is not adaptive for different datasets. Although ML with a fixed threshold achieves comparable results on PRW (29.2\% mAP and 79.0\% top-1), it performances poorly on CUHK-SYSU. Our dynamic threshold significantly alleviates this issue and achieves promising results on both datasets. 

\begin{figure*}[h]
  \centering
   \includegraphics[width=1\linewidth]{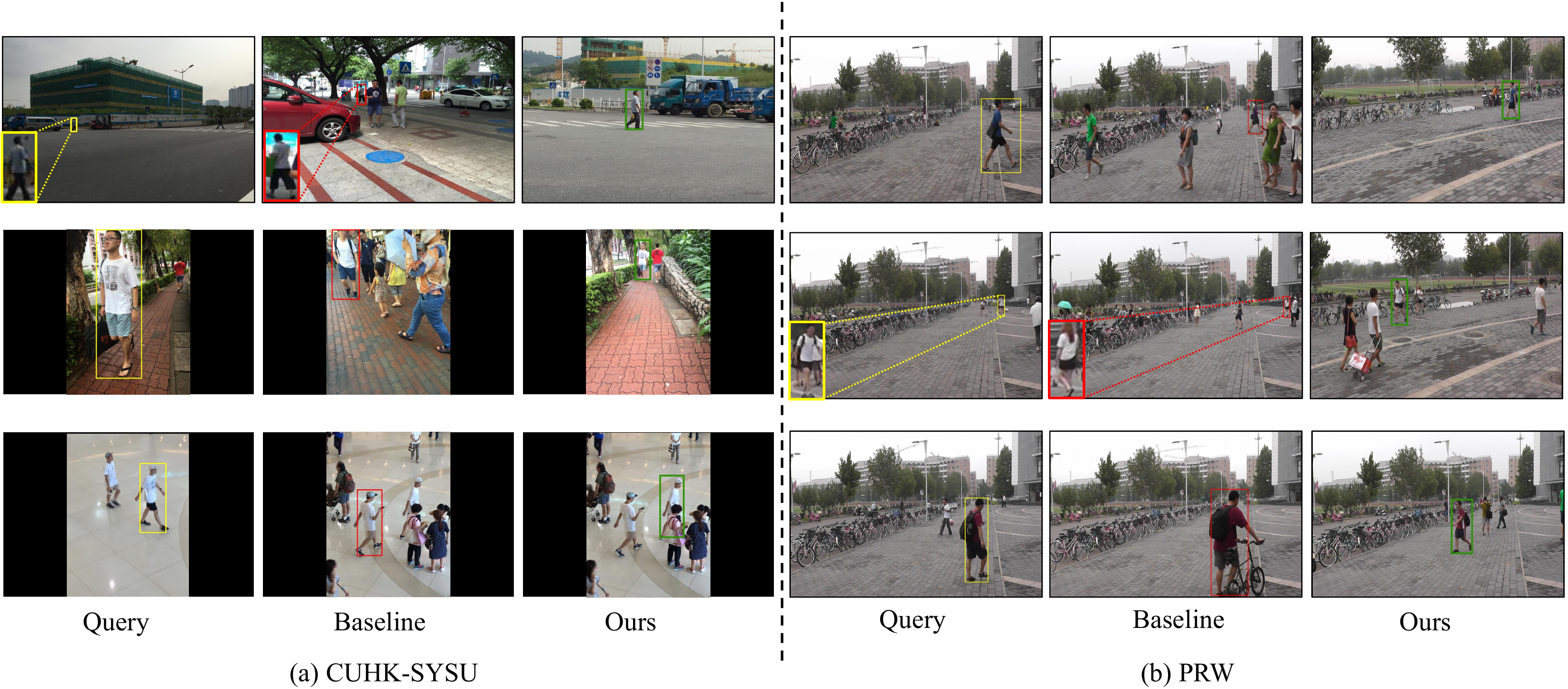}
   \caption{Rank-1 search results for several representative samples on CUHK-SYSU~\cite{xiao2017joint-OIM} and PRW~\cite{zheng2017person-PRW}. The green and red bounding boxes correspond to the correct and wrong results, respectively.}
   \label{fig:visualization}
\end{figure*}

\begin{figure}[H]
  \centering
   \includegraphics[width=0.7\linewidth]{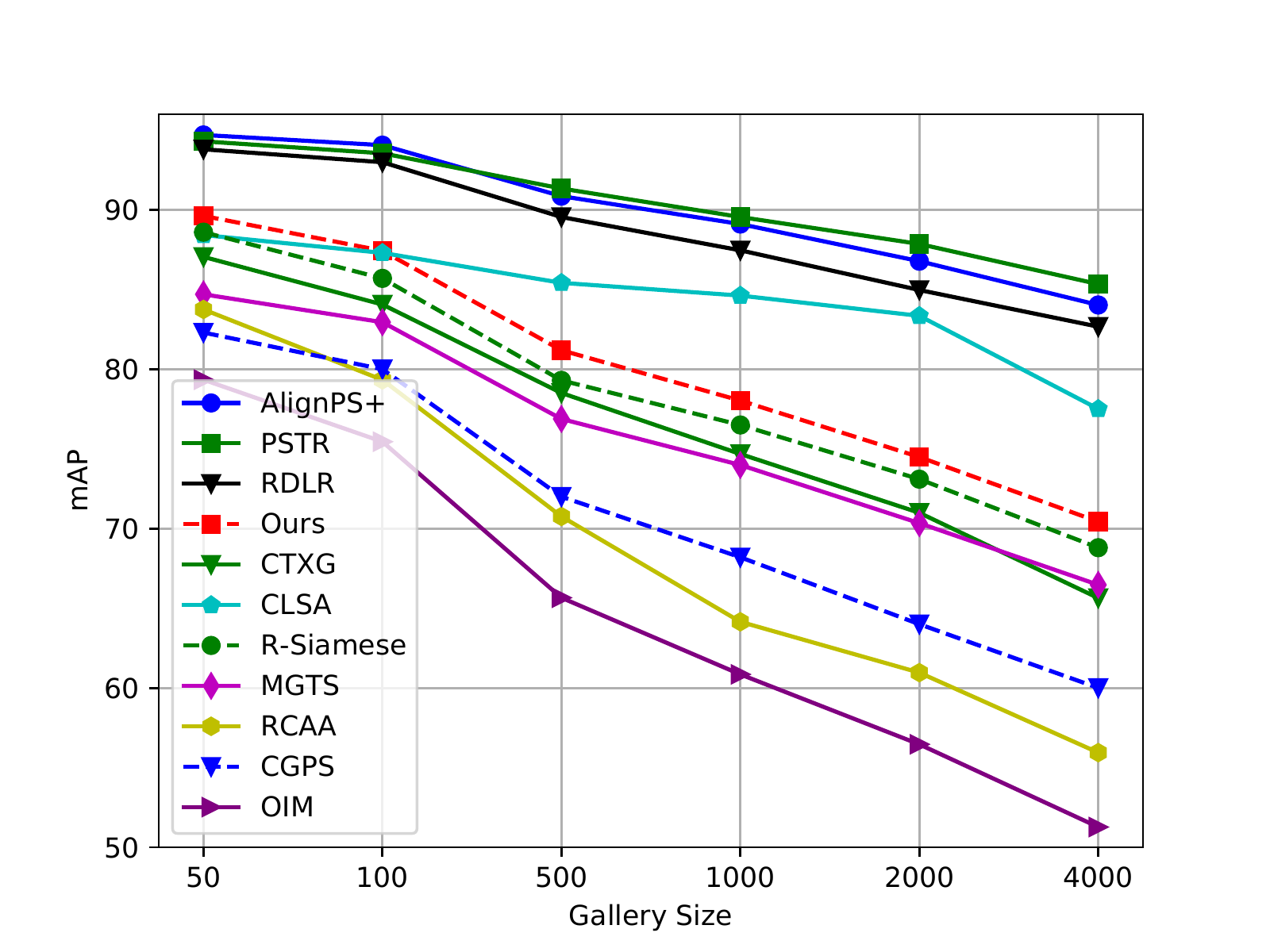}
   \caption{Performance comparison on the CUHK-SYSU dataset with different gallery size. The dashed lines denote the weakly supervised methods and the solid lines denote the fully supervised methods.}
   \label{fig:cuhk_gallery_size}
\end{figure}

\subsection{Comparison with the State-of-the-arts}
In this section, we compare our method with current state-of-the-art methods including fully supervised methods and weakly supervised methods.

\textbf{Results on CUHK-SYSU.}~\cref{tab:SOTA} shows the performance on CUHK-SYSU with the gallery size of 100. Our method achieves the best 87.4\% mAP and 88.5\% top-1, outperforming all existing weakly supervised person search methods. Specifically, we outperform the state-of-the-art method R-SiamNet by 1.4 in mAP\% and 1.4\% in top-1 accuracy. 
We also evaluate these methods under different gallery sizes from 50 to 4,000. In \cref{fig:cuhk_gallery_size}, we compare mAP with other methods. The dashed lines denote the weakly supervised methods and the solid lines denote the fully supervised methods. It can be observed that our method still outperforms all the weakly supervised with gallery increasing. Meanwhile, Our method surprisingly surpasses some fully supervised methods, e.g., \cite{xiao2017joint-OIM}, \cite{yan2019learning-CTXG},\cite{chang2018rcaa-RCAA} and \cite{chen2018person-mask-MGTS}. However, there still exists a significant performance gap. We hope our work could give some inspiration to others to explore weakly supervised person search. 
\begin{table}[b]
  \centering

    \begin{tabular}{l|l|cc|cc}
    \toprule
    \multicolumn{1}{c}{} & \multicolumn{1}{l}{\multirow{2}[4]{*}{Methods}} & \multicolumn{2}{c|}{ PRW } & \multicolumn{2}{c}{ CUHK-SYSU } \\
\cmidrule{3-6}    \multicolumn{1}{c}{} & \multicolumn{1}{c}{} & \multicolumn{1}{c}{ mAP } &  top-1  & \multicolumn{1}{c}{mAP} & top-1 \\
    \midrule
    \midrule
    \multirow{18}[2]{*}{\rotatebox{90}{Fully supervised}} 
          & OIM~\cite{xiao2017joint-OIM}   & 21.3  & 49.9  & 75.5  & 78.7 \\
          & IAN~\cite{xiao2019ian-IAN}   & 23.0    & 61.9  & 76.3  & 80.1 \\
          & NPSM~\cite{liu2017neural-NPSM}  & 24.2  & 53.1  & 77.9  & 81.2 \\
        &CTXG~\cite{yan2019learning-CTXG} & 33.4 & 73.6 & 84.1 & 86.5\\
          & MGTS~\cite{chen2018person-mask-MGTS}  & 32.6  & 72.1  & 83.0    & 83.7 \\
          & QEEPS~\cite{munjal2019query-QEEPS} & 37.1  & 76.7  & 88.9  & 89.1 \\
          & CLSA~\cite{lan2018person-multi-scale-matching-CLSA}  & 38.7  & 65.0    & 87.2  & 88.5 \\
          & HOIM~\cite{chen2020hierarchical-HOIM}  & 39.8  & 80.4  & 89.7  & 90.8 \\
          &APNet~\cite{zhong2020robust-APNet} & 41.9 & 81.4 & 88.9 & 89.3 \\
          
          & RDLR~\cite{han2019-RDLR}  & 42.9  & 70.2  & 93.0    & 94.2 \\
         & NAE~\cite{chen2020norm-NAE}   & 44.0    & 81.1  & 92.1  & 92.9 \\
        &PGS~\cite{kim2021prototype-PGS} & 44.2 & 85.2 & 92.3 & 94.7\\
          & BINet~\cite{dong2020bi-BINet} & 45.3  & 81.7  & 90.0    & 90.7 \\
            
          & AlignPS~\cite{yan2021anchor-AlignPS} & 45.9  & 81.9  & 93.1  & 93.4 \\
          & SeqNet~\cite{li2021sequential-SeqNet} & 46.7  & 83.4  & 93.8  & 94.6 \\
          & TCTS~\cite{wang2020tcts-TCTS}  & 46.8  & 87.5  & 93.9  & 95.1 \\
          & IGPN~\cite{dong2020instance-IGPN}  & 47.2  & 87.0    & 90.3  & 91.4 \\
          & OIMNet++~\cite{lee2022oimnet++} & 47.7  & 84.8  & 93.1  & 94.1 \\
          & PSTR~\cite{cao2022pstr-PSTR}  & 49.5  & 87.8  & 93.5  & 95.0 \\
          & AGWF~\cite{han2021end-AGWF}  & 53.3  & 87.7  & 93.3  & 94.2 \\
          & COAT~\cite{yu2022cascade-COAT}& 53.3  & 87.4  & 94.2  & 94.7 \\

    \midrule
    \midrule
    \multirow{3}[2]{*}{\rotatebox{90}{Weakly}} 
          & CGPS~\cite{yan2022exploring-CGPS}  & 16.2  & 68.0    & 80.0    & 82.3 \\
          & R-SiamNet~\cite{han2021weakly-R-SiamNet} & 21.2  & 73.4  & 86.0    & 87.1 \\
          \cline{2-6}
          
          & Ours  & \bf 30.7   & \bf 80.6   & \bf 87.4    & \bf 88.5 \\
    \bottomrule
    \end{tabular}%
    
      \caption{Comparison with the state-of-the-art methods on the PRW and CUHK-SYSU datasets. Weakly refers to the weakly supervised person search methods.}
  \label{tab:SOTA}%
\end{table}%

\textbf{Results on PRW.} As shown in \cref{tab:SOTA}, among existing two weakly supervised methods, CGPS~\cite{yan2022exploring-CGPS} and R-SiamNet~\cite{han2021weakly-R-SiamNet} achieve 16.2\%/68.0\% and 21.2\%/73.4\% in mAP/top-1. Our method achieves 30.7\%/80.6\% in mAP/top-1, surpassing all existing weakly supervised methods by a large margin. We argue that, as shown in~\cref{fig:scale-variation}, PRW has large variations of pedestrians scales, which presents multi-scale matching challenge, and our scale-invariant feature learning(\cref{sec: scale-invariant-feature-learning}) significantly alleviates this problem. As shown in \cref{tab:ablation study}, even the baseline model with our scale-invariant feature learning  still outperforms CGPS 10.0\%/8.4\% in mAP/top-1 and outperforms R-SiamNet in 5.0\%/3.0\% in mAP/top-1. 
\begin{figure}[!h]
  \centering
   \includegraphics[width=\linewidth]{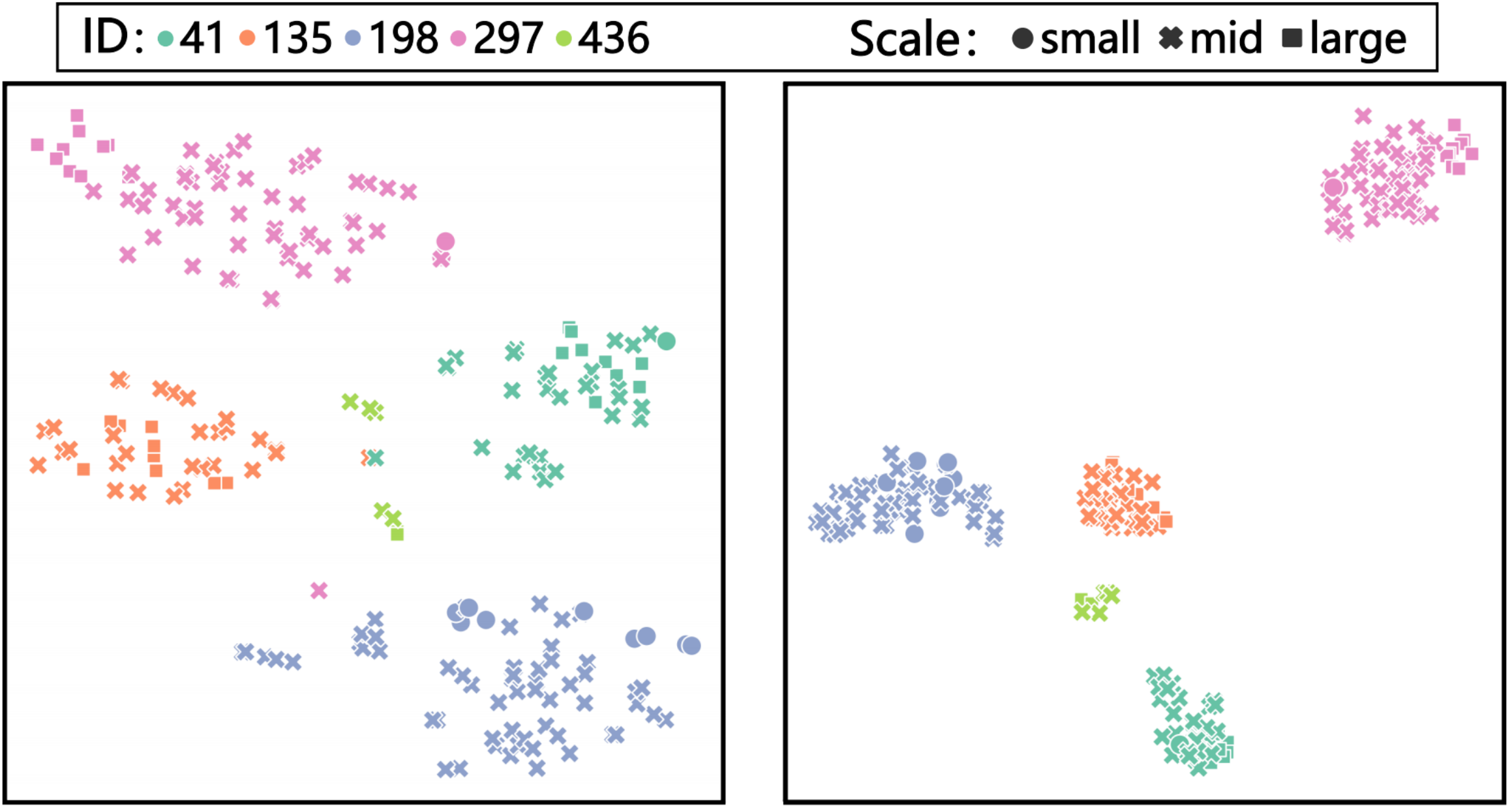}
   \caption{t-SNE feature visualization on part of the PRW training set. Colors denote different person identities and shapes denote persons at different scales.}
   \label{fig:tsne}
\end{figure}

\textbf{Visualization Analysis.} To evaluate the effectiveness of our method, we show several search results on CUHK-SYSU and PRW in~\cref{fig:visualization}. Specifically, the first two rows show that our method has stronger cross-scale retrieval capability compared to the baseline method. Additionally, the third row shows that our SSL extracts more discriminative features and retrieve the target person correctly among the confusing persons gallery.

Moreover, we visualize the feature distribution with t-SNE~\cite{van2008visualizing-tsne} in \cref{fig:tsne}. The circle denotes to the small scale persons whose resolution less than 3600 pixels, the square denotes to the large scale persons whose resolution larger than 45300, and the cross denotes to the medium scale persons whose resolution is between 3600 and 45300. Different colors represent different person identities. It illustrates that our method generates more consistent features across different scales. For more search results and visualizations, please refer to the supplementary materials.



\section{Conclusion}
In this paper, we propose a Self-similarity driven Scale-invariant Learning framework to solve the task of weakly supervised person search. With a scale-invariant loss, we can learn scale-invariant features by hard exemplar mining which will benefit the subsequent pseudo label prediction and person matching. We propose a dynamic multi-label learning method to generate pseudo labels and learn discrimative feature for re-id, which is adaptable to different datasets. To compensate dynamic multi-label learning, we also use cluster based strategy to learn re-id features. Finally, we learn the aforementioned parts in an end-to-end manner. Extensive experiments demonstrate that our proposed SSL can achieve state-of-the-art performance on two large-scale benchmarks.



\clearpage

{\small
\bibliographystyle{ieee_fullname}
\bibliography{egbib}
}

\end{document}